\documentclass[conference]{IEEEtran}
\IEEEoverridecommandlockouts
\usepackage{cite}
\usepackage{amsmath,amssymb,amsfonts}
\usepackage{algorithm2e}
\usepackage{algorithmic}
\usepackage{graphicx}
\usepackage{tabularx}
\graphicspath{{images/}{../images/}}
\usepackage{textcomp}
\usepackage{fvextra}
\usepackage{xcolor}
\usepackage{booktabs}
\usepackage{array}
\usepackage{multirow}
\usepackage{float}
\usepackage{tikz}
\usetikzlibrary{shapes.geometric, arrows.meta, positioning, calc, fit}
\usepackage[hidelinks]{hyperref}
\def\BibTeX{{\rm B\kern-.05em{\sc i\kern-.025em b}\kern-.08em
    T\kern-.1667em\lower.7ex\hbox{E}\kern-.125emX}}
\begin{document}
\sloppy
\hfuzz=10000pt
\hbadness=10000
\vbadness=10000
\overfullrule=0pt

\title{Characterizing Human-Likeness in AI Generated Poetry: A
Zero-shot Classification Study
}

\author{\scriptsize
\begin{tabular}{c@{\hspace{2.2em}}c@{\hspace{2.2em}}c}
\begin{tabular}[t]{@{}c@{}}
Angshu Nirmegh Biswas\\
\textit{Department of Computer Science and Engineering}\\
\textit{BRAC University}\\
angshu.nirmegh.biswas@g.bracu.ac.bd
\end{tabular}
&
\begin{tabular}[t]{@{}c@{}}
Taspiha Tabassum\\
\textit{Department of Computer Science and Engineering}\\
\textit{BRAC University}\\
taspiha.tabassum@g.bracu.ac.bd
\end{tabular}
&
\begin{tabular}[t]{@{}c@{}}
Afia Abida Shohid\\
\textit{Department of Computer Science and Engineering}\\
\textit{BRAC University}\\
afia.abida.shohid@g.bracu.ac.bd
\end{tabular}
\end{tabular}
\\[0.9em]
\begin{tabular}{c@{\hspace{4em}}c}
\begin{tabular}[t]{@{}c@{}}
Razia Marzan Mou\\
\textit{Department of Computer Science and Engineering}\\
\textit{BRAC University}\\
razia.marzan.mou@g.bracu.ac.bd
\end{tabular}
&
\begin{tabular}[t]{@{}c@{}}
Ayeeshah Akter Esha\\
\textit{Department of Computer Science and Engineering}\\
\textit{BRAC University}\\
ayeeshah.akter.esha@g.bracu.ac.bd
\end{tabular}
\end{tabular}
\\[0.9em]
\begin{tabular}{c@{\hspace{4em}}c}
\begin{tabular}[t]{@{}c@{}}
\textbf{Supervisor}\\
Dr. Farig Yousuf Sadeque\\
\textit{Associate Professor}\\
\textit{Department of Computer Science and Engineering}\\
\textit{BRAC University}
\end{tabular}
&
\begin{tabular}[t]{@{}c@{}}
\textbf{Co-supervisor}\\
Anika Ahmed\\
\textit{Lecturer}\\
\textit{Department of Computer Science and Engineering}\\
\textit{BRAC University}
\end{tabular}
\end{tabular}
}
\maketitle

\begin{abstract}
With the advancement of AI technologies, Generative AI (GenAI) and human written text have become nearly indistinguishable. Additionally, the global standardization of AI chatbots made academic malpractice more frequent. Furthermore, existing research indicates GenAI poems are the most difficult to distinguish even without any modification thus, GenAI poems are naturally deemed “human-like” by modern detectors. However, the objectivity of such dissertations needs to be verified against modern detection tools but the subjectivity of poetry and the “black-box” nature of the modern LLMs (Large Language Models) architectures made verification of such work quite complicated. Hence, the main objective of the research is to deduce the attributes of English poetry that contribute classification and misclassification of both human and AI poems and provide corroborating or contradicting evidence to the poetry distinguishability claim. For such characterizations, we propose a  Zero-shot detection pipeline with a dataset consisting of both human and AI poems to verify the distinguishability of human and AI creation and extract the aforementioned crucial attributes for accurate classification. Extraction of such attributes provides benefits in two ways: firstly, it reduces the margin of training needed as only the poems based on misclassifying attributes need to be trained and fine tuned and finally provides a critical insight to the GenAI detection dilemma to strengthen the modern detection pipelines.
\end{abstract}


\section{Introduction}
The use of GenAI has become very popular in academic writing and other fields resulting in academic malpractice and plagiarism which is especially true for the domain of the poetry. Poetry usually transcends the boundary of the grammatical structure and introduces ambiguity in forms of artistic expression. Such form of creativity blurs the line between human creativity and machine generated poems. The modern large language models show a strong ability in generating creative outputs, particularly poetry, by the use of metaphors, figurative language and stylistic variation which are traditionally associated with human creativity. Hence, surface detection metrics based on perplexity, burstiness, logarithmic probability (logprobs) \cite{5} and other forms of statistical analysis may struggle to accurately classify AI generation from human poems. Therefore, GenAI poetry detection requires extra depth, nuances and most importantly steady markers which we dubbed to be “Human-like” attributes to overcome the unreliable nature of statistics based classification. Moreover, the current detection tools are mostly trade secrets, which causes limited access for collaborative improvement. This is why, it is essential to develop open access systems that can not only identify the AI generated content but also provide room for improvement as collaboration will allow further human-like attributes and enhance the detection pipeline further by mending the underlying limitations.

\section{Related Works}
Difficulty of GenAI poetry classification is stated both through the naked eyes \cite{9} and the automated classification systems \cite{4}. This research aims to validate the aforementioned claims under one experimental setup and deduce how difficult it is to distinguish GenAI poems from human authored ones. Additionally, in poetry classification, MERMAID \cite{24} and Chen et al. \cite{22} highlighted the impacts of lexical diversity of GenAI poems and the usage of Metaphors as masking agents to humanize GenAI creation. Firstly, Chen et al. explains creative depth in human creation where human contents outrank models such as, GPT-2, GPT-Neo, LLaMA-2 and LLaMA-3 in lexical diversity and rhyming patterns. Secondly, per Chen et al., guided and structured based generation showed LLMs’ tendency to memorize and produce even less diverse output. However, MERMAID demonstrated using literary contents such as metaphors, the rigidness of GenAI contents can be broken. Therefore, just as MERMAID, this research aims to explore what other literary attributes blurs the line between human and machine generated content and answer whether structured generation (few-shot and style based prompting) can incorporate such literary attributes in GenAI content and moreover, what are the impacts of such attributes in the classification process as of whole. For this purpose, GPT OSS-120b \cite{28}, Qwen 3-32B \cite{30}, Llama 3.3-70B \cite{29} were selected to generate poems in a stylized and human few-shot setting. Finally, the combination both of human samples, generated creation of the aforementioned models were passed on to classify through Gemma 4-31B \cite{36} (in a Zero-Shot environment) and to human survey participants to affirm the claims of RAID benchmark and Porter and Machery and record the “Humanizing” and the “Non-humanizaing” attributes to strengthen the classification.

\section{Methodology Overview}
The research methodology is divided into three branches: dataset creation, classification, and analysis.

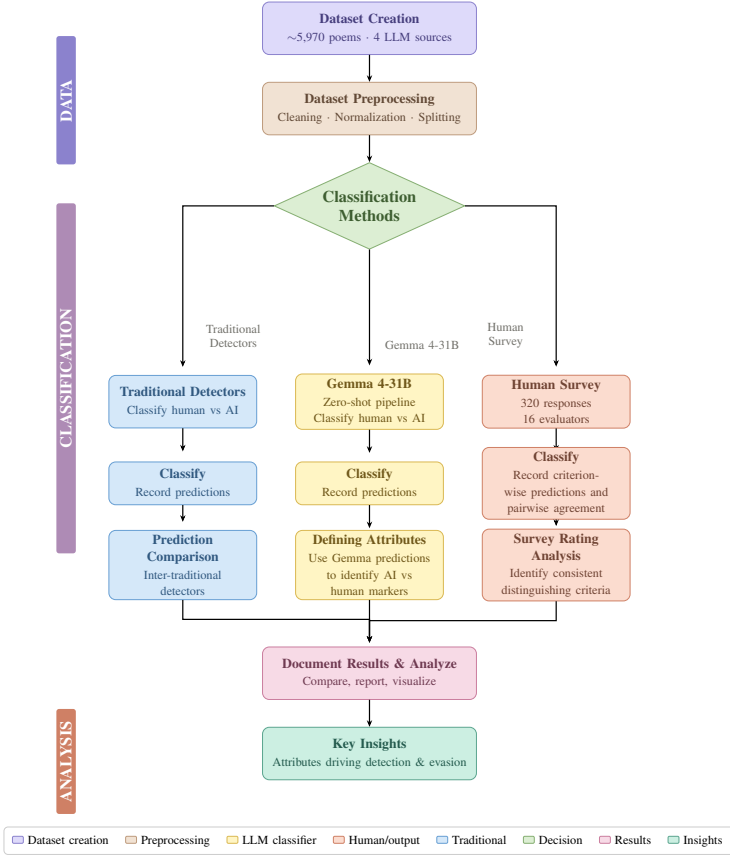
\begin{figure}[H]
  \centering
  \makebox[\columnwidth][l]{\hspace*{-0.35\columnwidth}\resizebox{1.35\columnwidth}{!}{
\definecolor{purplefill}{RGB}{222,218,250}
\definecolor{purplestroke}{RGB}{139,130,205}
\definecolor{purpletext}{RGB}{70,62,125}
\definecolor{tealfill}{RGB}{255,245,196}
\definecolor{tealstroke}{RGB}{218,184,82}
\definecolor{tealtext}{RGB}{112,88,26}
\definecolor{coralfill}{RGB}{251,220,207}
\definecolor{coralstroke}{RGB}{207,128,101}
\definecolor{coraltext}{RGB}{112,68,52}
\definecolor{bluefill}{RGB}{213,232,249}
\definecolor{bluestroke}{RGB}{102,158,211}
\definecolor{bluetext}{RGB}{49,89,128}
\definecolor{amberfill}{RGB}{252,231,188}
\definecolor{amberstroke}{RGB}{204,154,72}
\definecolor{ambertext}{RGB}{113,82,36}
\definecolor{greendark}{RGB}{197,232,220}
\definecolor{greenstroke}{RGB}{91,165,143}
\definecolor{teallight}{RGB}{36,91,78}
\definecolor{arrowgray}{RGB}{166,166,160}
\definecolor{labelgray}{RGB}{115,114,108}
\definecolor{sagefill}{RGB}{220,238,211}
\definecolor{sagestroke}{RGB}{133,174,113}
\definecolor{sagetext}{RGB}{62,94,49}
\definecolor{rosefill}{RGB}{247,218,230}
\definecolor{rosestroke}{RGB}{203,126,159}
\definecolor{rosetext}{RGB}{111,58,81}
\definecolor{skyfill}{RGB}{204,239,226}
\definecolor{skystroke}{RGB}{91,165,143}
\definecolor{skytext}{RGB}{36,91,78}
\definecolor{nudefill}{RGB}{241,226,211}
\definecolor{nudestroke}{RGB}{190,155,125}
\definecolor{nudetext}{RGB}{98,73,54}
\definecolor{phaseclassfill}{RGB}{174,139,181}
\definecolor{phaseclasstext}{RGB}{255,255,255}

\tikzset{
  basenode/.style={
    rectangle, rounded corners=5pt,
    minimum width=5.5cm, minimum height=1.35cm,
    align=center, text width=5cm,
    font=\normalsize,
    line width=0.4pt
  },
  purplenode/.style={basenode,
    fill=purplefill, draw=purplestroke, text=purpletext},
  tealnode/.style={basenode,
    fill=tealfill, draw=tealstroke, text=tealtext},
  coralnode/.style={basenode,
    fill=coralfill, draw=coralstroke, text=coraltext},
  bluenode/.style={basenode,
    fill=bluefill, draw=bluestroke, text=bluetext},
  ambernode/.style={basenode,
    fill=amberfill, draw=amberstroke, text=ambertext},
  smallnode/.style={basenode,
    minimum width=3.8cm, text width=3.5cm},
  tinynode/.style={basenode,
    minimum width=3.8cm, minimum height=1.1cm, text width=3.5cm},
  diamondnode/.style={
    diamond, aspect=2.2,
    minimum width=4cm, minimum height=1.6cm,
    align=center, fill=sagefill, draw=sagestroke,
    text=sagetext, font=\large\bfseries, line width=0.4pt
  },
  myarrow/.style={
    -{Stealth[length=5pt,width=4pt]},
    draw=black, line width=0.9pt
  },
  phaselabel/.style={
    rotate=90, font=\large\bfseries,
    anchor=center
  }
}

\begin{tikzpicture}[node distance=0.7cm, xshift=0.45cm]

\newcommand{\xcenter}{0}
\newcommand{\xleft}{-4.8}
\newcommand{\xright}{4.8}

\fill[purplestroke, rounded corners=2pt]
  (-8.05,-0.2) rectangle (-7.55,-3.5);
\node[phaselabel, text=white] at (-7.8,-1.85) {DATA};

\fill[phaseclassfill, rounded corners=2pt]
  (-8.05,-4.5) rectangle (-7.55,-13.5);
\node[phaselabel, text=phaseclasstext] at (-7.8,-9.0) {CLASSIFICATION};

\fill[coralstroke, rounded corners=2pt]
  (-8.05,-17.5) rectangle (-7.55,-20.2);
\node[phaselabel, text=white] at (-7.8,-18.85) {ANALYSIS};

\node[purplenode] (creation) at (\xcenter, 0) {
  \textbf{Dataset Creation}\\[1pt]
  {\small $\sim$5,970 poems $\cdot$ 4 LLM sources}
};

\draw[myarrow] (creation.south) -- ++(0,-0.7)
  node[midway] {} coordinate (a1);

\node[basenode, fill=nudefill, draw=nudestroke, text=nudetext, below=0.7cm of creation] (preproc) {
  \textbf{Dataset Preprocessing}\\[1pt]
  {\small Cleaning $\cdot$ Normalization $\cdot$ Splitting}
};

\draw[myarrow] (preproc.south) -- ++(0,-0.7);

\node[diamondnode, below=0.7cm of preproc] (diamond) {
  Classification\\Methods
};


\node[font=\small, text=labelgray, align=center]
  at (-3.5,-7.9) {Traditional\\Detectors};
\node[font=\small, text=labelgray, align=center, anchor=west]
  at (0.28,-8.15) {Gemma 4-31B};
\node[font=\small, text=labelgray, align=center]
  at (3.5,-7.9) {Human\\Survey};

\node[bluenode, smallnode] (trad) at (\xleft,-9.6) {
  \textbf{Traditional Detectors}\\[1pt]
  {\small Classify human vs AI}
};

\draw[myarrow]
  (diamond.west) -- (trad |- diamond.west) -- ([yshift=6pt]trad.north);

\node[tealnode, smallnode] (gemma) at (\xcenter,-9.6) {
  \textbf{Gemma 4-31B}\\[1pt]
  {\small Zero-shot pipeline}\\
  {\small Classify human vs AI}
};

\node[coralnode, smallnode] (survey) at (\xright,-9.6) {
  \textbf{Human Survey}\\[1pt]
  {\small 320 responses}\\
  {\small 16 evaluators}
};

\draw[myarrow]
  (diamond.east) -- (survey |- diamond.east) -- ([yshift=6pt]survey.north);

\draw[myarrow] (diamond.south) -- ([yshift=6pt]gemma.north);

\draw[myarrow] (trad.south) -- ++(0,-0.6);
\draw[myarrow] (gemma.south) -- ++(0,-0.6);
\draw[myarrow] (survey.south) -- ++(0,-0.6);

\node[bluenode, tinynode] (ctrad) at (\xleft,-11.7) {
  \textbf{Classify}\\[1pt]
  {\small Record predictions}
};

\node[tealnode, tinynode] (cgemma) at (\xcenter,-11.7) {
  \textbf{Classify}\\[1pt]
  {\small Record predictions}
};

\node[coralnode, tinynode] (csurvey) at (\xright,-11.7) {
  \textbf{Classify}\\[1pt]
  {\small Record criterion-wise predictions and pairwise agreement}
};

\draw[myarrow] (cgemma.south) -- ++(0,-0.6);

\node[bluenode, smallnode] (atrad) at (\xleft,-13.8) {
  \textbf{Prediction Comparison}\\[1pt]
  {\small Inter-traditional}\\
  {\small detectors}
};

\draw[myarrow] (ctrad.south) -- ([yshift=6pt]atrad.north);

\node[tealnode, smallnode] (agemma) at (\xcenter,-13.8) {
  \textbf{Defining Attributes}\\[1pt]
  {\small Use Gemma predictions}\\
  {\small to identify AI vs human markers}
};

\node[coralnode, smallnode] (asurvey) at (\xright,-13.8) {
  \textbf{Survey Rating Analysis}\\[1pt]
  {\small Identify consistent distinguishing criteria}
};

\draw[myarrow, shorten >=2pt]
  (csurvey.south) -- (asurvey.north);

\coordinate (merge) at (0,-15.8);

\draw[myarrow]
  (atrad.south) -- ++(0,-0.5) -| (merge);
\draw[myarrow]
  (agemma.south) -- (merge);
\draw[myarrow]
  (asurvey.south) -- ++(0,-0.5) -| (merge);

\node[basenode, fill=rosefill, draw=rosestroke, text=rosetext, below=0.1cm] (docresults) at (merge) {
  \textbf{Document Results \& Analyze}\\[1pt]
  {\small Compare, report, visualize}
};

\draw[myarrow] (docresults.south) -- ++(0,-0.7);

\node[basenode, fill=skyfill, draw=skystroke, text=skytext, below=0.7cm of docresults] (insights) {
  \textbf{Key Insights}\\[1pt]
  {\small Attributes driving detection \& evasion}
};


\node[draw=gray!40, rounded corners=3pt, line width=0.4pt,
      inner sep=6pt, below=1.2cm of insights,
      font=\small, align=left] (legend) {
  \tikz\fill[purplefill, draw=purplestroke, rounded corners=1pt]
    (0,0) rectangle (0.28,0.18); Dataset creation \quad
  \tikz\fill[nudefill, draw=nudestroke, rounded corners=1pt]
    (0,0) rectangle (0.28,0.18); Preprocessing \quad
  \tikz\fill[tealfill, draw=tealstroke, rounded corners=1pt]
    (0,0) rectangle (0.28,0.18); LLM classifier \quad
  \tikz\fill[coralfill, draw=coralstroke, rounded corners=1pt]
    (0,0) rectangle (0.28,0.18); Human/output \quad
  \tikz\fill[bluefill, draw=bluestroke, rounded corners=1pt]
    (0,0) rectangle (0.28,0.18); Traditional \quad
  \tikz\fill[sagefill, draw=sagestroke, rounded corners=1pt]
    (0,0) rectangle (0.28,0.18); Decision \quad
  \tikz\fill[rosefill, draw=rosestroke, rounded corners=1pt]
    (0,0) rectangle (0.28,0.18); Results \quad
  \tikz\fill[skyfill, draw=skystroke, rounded corners=1pt]
    (0,0) rectangle (0.28,0.18); Insights
};

\end{tikzpicture}}}
  \caption{Overview of the proposed methodology for detecting AI-generated poetry.}
  \label{fig:methodology}
\end{figure}

The elaboration of each branch is described in the following subsections.

\subsection{Dataset Generation Pipeline in Detail}
\begin{figure}[!t]
    \centering
    \includegraphics[width=\columnwidth]{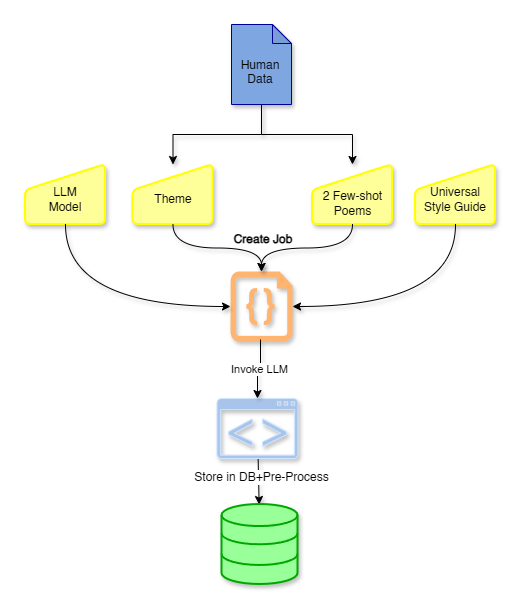}
    \caption{Data generation pipeline}
    \label{fig:data-generation-pipeline}
\end{figure}

The workflow shown in Fig.~\ref{fig:data-generation-pipeline} represents how data samples are generated by LLMs based on the human sample containing human-written poems. In this section, the generation pipeline is described in detail.

\subsubsection{Human Sample Collection}\label{sec:4.1.1}
Human-written poems were gathered from three sources: poetry posted publicly on social media platforms, personal archives shared directly by authors after permission was requested, and open-access literary websites where authors publish their poems. We approached the authors individually for permission to work with their poems, and 1,513 human poems were collected through this process, including 11 authors and diverse writing styles.

\subsubsection{Initialization and LLM Generation Loop}
For poem generation, three LLMs were selected based on novelty: GPT-OSS-120B, Llama 3.3-70B, and Qwen 3-32B. Before invoking the LLMs, each incomplete generation request was packaged as a job. This package included a predetermined version of the Universal Style Guide and a randomly selected theme a global theme array containing 20 traditional poetry themes. The job also included two few-shot examples based on the author style being mimicked. The purpose of the few-shot style examples was to help the models understand and learn the author's style without memorizing or copying the work verbatim. This job packing framework is much lighter than Wang et al.'s \cite{38} image based poetry generation ableit at the cost less sophisticated generation. Finally, the jobs were shuffled and sent to the LLMs for generation.

\paragraph{Universal Style Guide}\label{sec:4.1.1.2.1}
\begin{itemize}
    \item \textbf{Titling:} A meaningful, original title must precede the generated text.
    \item \textbf{Linguistic Constraints:} Outputs are strictly limited to English, and the use of em dashes (---) is explicitly prohibited.
    \item \textbf{Originality:} The model must mimic the stylistic voice of the provided examples without plagiarizing content, ensuring a novel generation based on the assigned theme.
\end{itemize}

\paragraph{Global Thematic Pool}
\begin{equation}
    T = \{\text{Love}, \text{Romanticism}, \text{Heartbreak}, \dots, \text{Wonder}, \text{Resilience}\}
\end{equation}
Here, $T$ represents the global theme array, which was later packed as part of each job.

\paragraph{Dynamic Few-Shot Construction}
\begin{equation}
    S_a = \text{RandomSample}(D_a, \min(n, |D_a|))
\end{equation}
\begin{equation}\label{eq:few-shot-prompt}
    P_{FS}(a) = I_{prefix} \oplus \bigoplus_{s \in S_a} \Big(\text{Title}(s) \oplus \text{Poem}(s)\Big)
\end{equation}
Here, $S_a$ is the subset of sampled few-shot examples for target author $a$. $D_a$ is the subset of the human dataset containing only works authored by $a$, and $|D_a|$ is the total number of available works for that author. $n$ is the target number of few-shot examples to include, which was 2 in this study. $P_{FS}(a)$ is the final constructed few-shot prompt for author $a$ (See Example:~\ref{app:prompt_example}). $I_{prefix}$ represents the initial system instruction block, or Universal Style Guide. $s$ is an individual sampled record from which $\text{Title}(s)$ and $\text{Poem}(s)$ are extracted as strings.

\paragraph{Job Queue Creation}
\begin{center}
\begin{minipage}{\columnwidth}
\refstepcounter{algocf}\label{alg:job-queue-formulation}
\hrule\smallskip
\noindent\textbf{Algorithm~\thealgocf: Job Queue Formulation with Resume Logic}
\smallskip\hrule\smallskip
\footnotesize
\begin{algorithmic}[1]
\REQUIRE Human dataset $D$, model set $M$, theme set $T$, and completed jobs $C$
\ENSURE Randomized pending task queue $J$
\STATE $J \leftarrow \emptyset$
\FOR{each $(index, entry) \in D$}
    \STATE $author \leftarrow entry.\text{Author}$
    \FOR{each $model \in M$}
        \IF{$(index, model) \notin C$}
            \STATE $theme \leftarrow \text{RandomSample}(T)$
            \STATE $J \leftarrow J \cup \{(model, author, theme, index)\}$
        \ENDIF
    \ENDFOR
\ENDFOR
\RETURN $\text{Shuffle}(J)$
\end{algorithmic}
\smallskip\hrule
\end{minipage}
\end{center}
From the incomplete generation tasks, queue $J$ is created. The ledger of completed tasks, $C$, and the value $index$ are kept to track progress and make the overall process crash-consistent during API calls.

\paragraph{LLM Invocation}
\begin{equation}
    O_{raw} = f_m(P_{FS}(a) \oplus P_{user}(t))
\end{equation}
Here, $O_{raw}$ is the raw generated output string returned by the language model. $f_m$ is the function representing the LLM, where $m \in M$. $P_{FS}(a)$ is the complete style guide constructed in Eq.~\eqref{eq:few-shot-prompt}. $P_{user}(t)$ is the explicit user prompt instructing the model to generate a text based on the designated theme $t \in T$.

\subsection{Dataset Overview and Preprocessing}\label{sec:4.1.2}
The final dataset contains 5,970 poems, including 4,539 AI-generated poems and 1,513 human-authored poems which is signifiacntly more than the POEMetric corpus\cite{37}. AI poems were generated using Llama 3.3-70B, GPT-OSS-120B, and Qwen 3-32B. All poems were generated under the few-shot prompt setting. Each model generated 1,513 poems, and each human poem served as a style reference during few-shot generation. From a pool of 20 universal themes, each theme was assigned randomly.

\subsubsection{Data Preprocessing}
A cleaning pipeline was applied to preprocess raw outputs from AI-generated poems. First, chain-of-thought blocks enclosed in \texttt{<think>} were removed in cases where the model failed to close the tag after reaching token limits. Second, leftover generation code fences and language identifiers were removed. Third, phrases such as ``Here is a poem'' and explicit \texttt{Title:} or \texttt{Theme:} headers were stripped. Typographic and symbolic noise, including em dashes, Unicode dash variants, and double-hyphen variants, was removed from line endings and replaced with spaces elsewhere. Standard punctuation was retained, while mathematical operators, brackets, and special symbols were removed. Finally, triple dots and ellipses were normalized. Because this study focuses on English poems, non-English words generated by Qwen inside English poems were filtered out. Whitespace was standardized, and consecutive blank lines were reduced to one empty line between stanzas. From the 4,539 AI samples, 82 poems were removed during cleaning because of poor generation quality.

\begin{figure}[!t]
    \centering
    \includegraphics[width=\columnwidth]{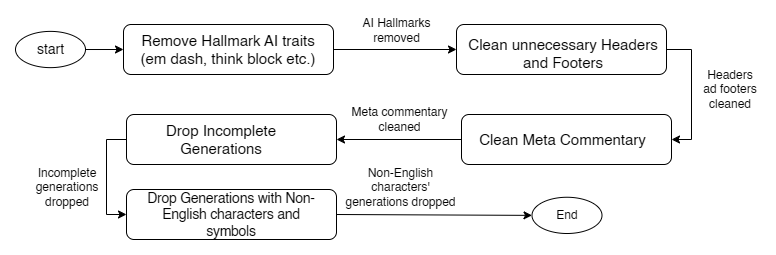}
    \caption{Data cleaning visualization}
    \label{fig:data-cleaning}
\end{figure}

\subsubsection{Feature Extraction}
To compare the differences between AI-generated and human poems, vocabulary- and phrase-based features were extracted from the dataset. The analysis included total word count, unique words, type-token ratio (TTR), most frequent words, bigrams, trigrams, semantic consistency, prompt repetition, vocabulary overlap, and TF-IDF. The summary of the Feature Extraction methods can be found at Table ~\ref{tab:extracted-features}. These features were used to determine whether AI-generated poems rely on the semantics produced by human authors.

\begin{table}[!t]
\centering
\caption{Extracted vocabulary and n-gram features}
\label{tab:extracted-features}
\footnotesize
\setlength{\tabcolsep}{2pt}
\renewcommand{\arraystretch}{1.35}
\begin{tabular}{p{0.20\columnwidth}p{0.34\columnwidth}p{0.36\columnwidth}}
\toprule
\textbf{Feature} & \textbf{Definition} & \textbf{Purpose} \\
\midrule
Total words & Sum of all word tokens after preprocessing & Establishes corpus size and enables normalized comparison. \\
Unique words & Count of distinct word types & Measures vocabulary richness and lexical repetition. \\
Type-token ratio (TTR) & Unique words divided by total words & Measures lexical diversity; higher TTR indicates less lexical repetition. \\
Top frequent words & Most frequent words in each corpus & Identifies dominant vocabulary and repeated poetic terms. \\
Bigrams & Consecutive two-word phrases & Captures repeated two-word phrase patterns. \\
Trigrams & Consecutive three-word phrases & Captures repeated three-word phrase patterns. \\
TF-IDF & Words that are most representative of each source & Shows source-specific vocabulary. \\
Semantic consistency & Theme leakage consistency score measured from assigned theme words and other theme words & Checks whether generated poems remain focused on the assigned theme. \\
Prompt repetition & Frequency and presence of assigned theme words inside generated poems & Detects whether the model repeats the assigned theme word directly. \\
Vocabulary overlap & Shared words, bigrams, trigrams, and bigram overlap frequency & Measures how much AI vocabulary and phrase construction overlap with human poems. \\
\bottomrule
\end{tabular}
\end{table}

\subsubsection{Theme-Based Dataset and Distribution}
From the AI-generated poems across 20 themes, theme analysis was performed to examine whether vocabulary patterns change when poems are grouped by theme.

\begin{table}[!t]
\centering
\caption{Poem count per theme and model}
\label{tab:theme-model-counts}
\resizebox{\columnwidth}{!}{%
\begin{tabular}{lrrrr}
\toprule
\textbf{Theme} & \textbf{GPT-OSS-120B} & \textbf{Llama-3.3-70B} & \textbf{Qwen-3-32B} & \textbf{Total} \\
\midrule
Beauty & 65 & 76 & 71 & 212 \\
Betrayal & 66 & 80 & 78 & 224 \\
Desire & 73 & 72 & 63 & 208 \\
Eternity & 72 & 82 & 85 & 239 \\
Faith & 68 & 68 & 74 & 210 \\
Grief & 63 & 81 & 65 & 209 \\
Heartbreak & 85 & 80 & 82 & 247 \\
Hope & 83 & 77 & 53 & 213 \\
Joy & 72 & 91 & 80 & 243 \\
Love & 68 & 65 & 71 & 204 \\
Melancholy & 75 & 76 & 77 & 228 \\
Mortality & 73 & 60 & 68 & 201 \\
Nature & 77 & 81 & 75 & 233 \\
Nostalgia & 83 & 86 & 75 & 244 \\
Rebirth & 72 & 69 & 79 & 220 \\
Resilience & 73 & 76 & 66 & 215 \\
Romanticism & 83 & 64 & 75 & 222 \\
Solitude & 78 & 77 & 75 & 230 \\
Sorrow & 85 & 88 & 69 & 242 \\
Wonder & 84 & 59 & 70 & 213 \\
\bottomrule
\end{tabular}}
\end{table}

\begin{figure}[!t]
    \centering
    \includegraphics[width=\columnwidth]{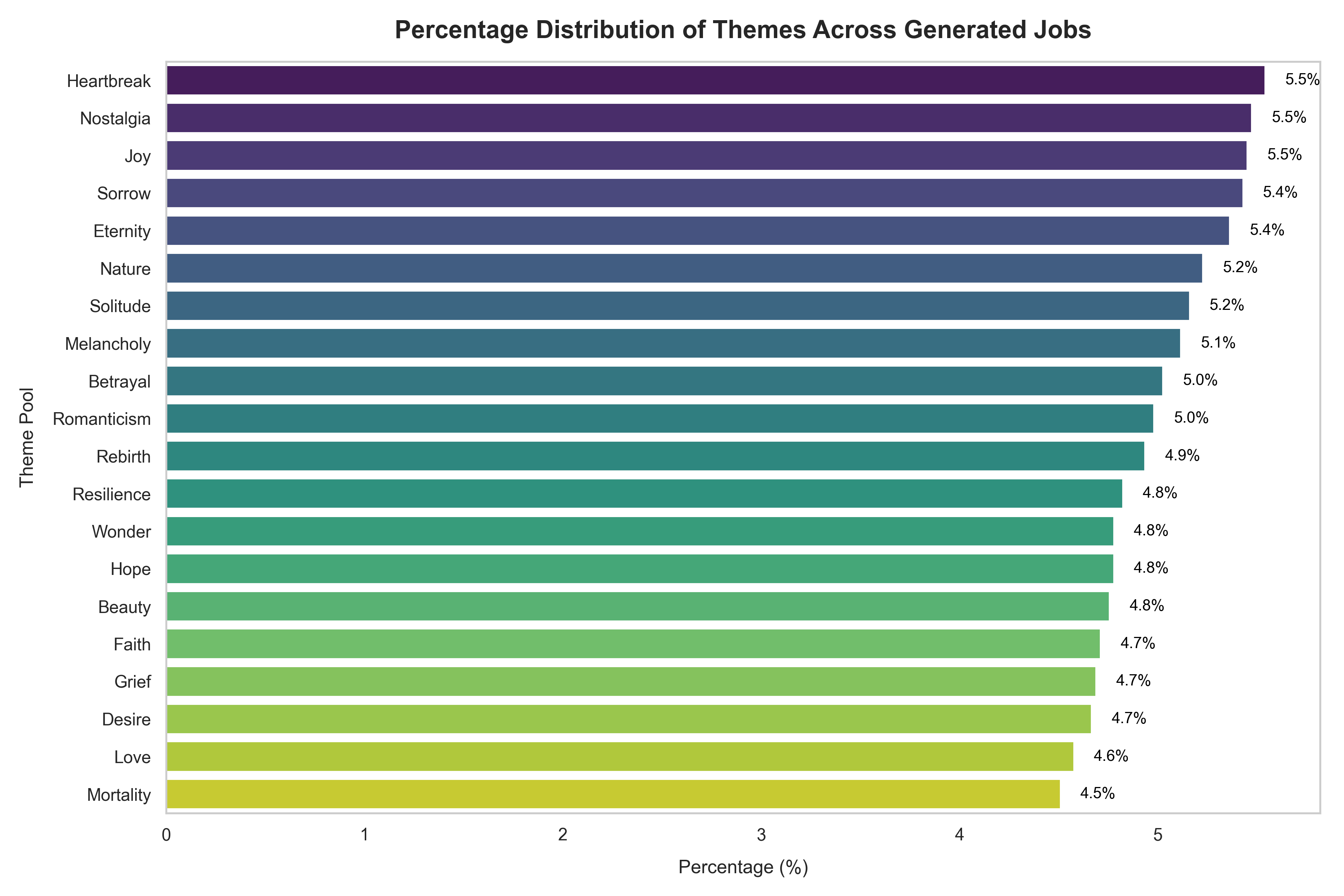}
    \caption{Theme percentage distribution across all LLM invocation calls}
    \label{fig:theme-percentage-distribution}
\end{figure}

The theme distribution is almost balanced across the models, with the highest difference between the themes Heartbreak and Mortality being approximately 1\% (Fig.~\ref{fig:theme-percentage-distribution}) and Table ~\ref{tab:theme-model-counts} displays theme distribution per model is also near equal.  

\subsection{Human Classification via Survey}\label{sec:4.1.4}
\subsubsection{Custom Survey System}
We built a web application to collect survey responses from humans. This website served as a platform for collecting human evaluations to understand perceptions of poems in our custom dataset. The primary goal was to gather detailed judgments from users who analyzed poems according to different metrics and decided whether each poem appeared to be AI-generated or human-written. Every participant was presented with a set of twenty poems and asked to evaluate them using a consistent and structured approach.

The evaluation process used five criteria: clarity of expression, use of literary devices, punctuation and spacing patterns, grammar and spelling accuracy, and overall originality. For each criterion, participants rated how likely they thought the poem was AI-generated using a Likert scale ranging from ``Very Unlikely'' to ``Very Likely''. The platform also offered an optional comment section where evaluators could describe specific elements that influenced their decisions.

The website was designed for both user convenience and research integrity. Participants began by providing an email address, which acted as a unique identifier and helped track progress. A progress bar displayed how many poems the user had evaluated out of twenty. The system saved progress so that participants could resume the survey later. The system also assigned poems in a way that maintained fair distribution across users while preserving research validity.

\subsection{Gemma 4 Zero-Shot Classification}\label{sec:4.1.6}
Gemma 4-31B was used as a filter to identify patterns exhibited by GenAI poems. Previous research has shown that GenAI poetry can be difficult to distinguish from human poetry and may even be preferred over human-written poems. Therefore, to extract telltale AI patterns, Gemma 4 was used in a zero-shot setting, meaning it had no prior knowledge or training on our dataset of 5,970 samples. This method also helped identify which machine-generated attributes appeared human-like. These machine-like and human-like attributes produced hard positive results, which we call defining attributes, and hard negative results, which are attributes that force misclassification.

\subsubsection{Ensuring the Zero-Shot Setting}
To ensure the zero-shot setting, Gemma 4 was queried with a single random line from human poems and asked to return the next line. Gemma was queried over 100 human samples from 11 human authors. If Gemma returned the next poetry line correctly, that author's work would be considered contaminated and discarded from the human dataset.

\subsubsection{Classification}
The classification process required the dataset to be shuffled to reduce unpredictable ordering biases. Each sample was queried to be classified using the same five criteria and Likert scale as the survey, with three additional responses: classification label, confidence score from 0--100, and reasoning factor. The reasoning factor asked which of the five criteria most influenced the label, or requested another factor if none of the five applied.

\begin{center}
\begin{minipage}{\columnwidth}
\refstepcounter{algocf}\label{alg:gemma4_detection_pipeline}
\hrule\smallskip
\noindent\textbf{Algorithm~\thealgocf: Gemma 4 Detection Classification Pipeline}
\smallskip\hrule\smallskip
\footnotesize
\textbf{Input:} Raw dataset $\mathcal{F}_{\text{in}}$, checkpoint $\mathcal{F}_{\text{out}}$, and criteria prompt $\mathcal{S}_{\text{prompt}}$\\
\textbf{Output:} Evaluated DataFrame continuously committed to $\mathcal{F}_{\text{out}}$
\begin{algorithmic}[1]
\STATE Load $\mathcal{DF}$ from $\mathcal{F}_{\text{out}}$ if it exists; otherwise load $\mathcal{F}_{\text{in}}$
\STATE Initialize missing detection labels, confidence, reasoning, five metric columns, and processed flags
\STATE Cast $\mathcal{DF}[\text{Processed}]$ to Boolean and set metric columns to object datatypes
\FOR{each row index $i$ and record $R$ in $\mathcal{DF}$}
    \IF{$R[\text{Processed}]$ is true}
        \STATE continue to next record
    \ENDIF
    \STATE $poem\_text \leftarrow R[\text{poem}]$
    \IF{$poem\_text$ is null or empty}
        \STATE mark $R$ processed and continue
    \ENDIF
    \STATE $\mathcal{S}_{\text{flag}} \leftarrow \text{False}$
    \WHILE{$\mathcal{S}_{\text{flag}}$ is false}
        \IF{Gemma API returns valid $raw\_content$}
            \STATE $\mathcal{J} \leftarrow \text{Extract\_JSON\_From\_Text}(raw\_content)$
            \IF{$\mathcal{J}$ is a valid dictionary}
                \STATE Copy label, confidence, reasoning, five metric scores, and processed flag from $\mathcal{J}$ to $R$
            \ELSE
                \STATE Store raw output in $R[\text{Reasoning}]$ and mark $R$ processed
            \ENDIF
        \ELSE
            \STATE Store empty-response error in $R[\text{Reasoning}]$ and mark $R$ processed
        \ENDIF
        \STATE Save $\mathcal{DF}$ to $\mathcal{F}_{\text{out}}$, set $\mathcal{S}_{\text{flag}}$ true, and sleep for 2 seconds
        \STATE On network exception, log the error and sleep for 60 seconds
    \ENDWHILE
\ENDFOR
\end{algorithmic}
\smallskip\hrule
\end{minipage}
\end{center}

The $\mathcal{S}_{flag}$ flag and $R[\text{Processed}]$ record state were used to prevent API token overuse and make the system crash-consistent.

\subsection{Traditional Detector Classification}\label{sec:4.1.7}
Apart from Gemma 4, four additional detectors were used to evaluate the dataset as comparison metrics: Log-Likelihood~\cite{34}, Log-Rank Ratio (LRR)~\cite{35}, Binoculars~\cite{33}, and Fast-DetectGPT~\cite{32}. These detectors are pretrained and can efficiently classify unaltered GenAI content. Testing our dataset against these models provides insight into Gemma 4's classification capability and the evasiveness of the dataset.

\section{Results}
The results are organized around lexical diversity, n-gram overlap, survey-based human evaluation, Gemma classification, traditional detector performance, and qualitative attributes of both GenAI and human-authored poems.

\subsection{Word Diversity Analysis}
GPT-OSS-120B, Llama-3.3-70B, and Qwen-3-32B were compared with human samples using the methods explained in ~\ref{tab:extracted-features}. The following subsections elaborate on the word diversity analysis.

\subsubsection{Type-Token Ratio and TF-IDF Analysis}
TTR was calculated for each corpus by dividing the number of unique word types by the total number of word tokens after removing stopwords. The human corpus achieved the highest TTR, indicating the richest vocabulary relative to corpus size. Among the AI models, Qwen achieved the highest TTR, while GPT and Llama showed lower word variation and higher repetition.

\begin{table}[H]
\centering
\caption{Total words, unique words, and TTR by source}
\label{tab:ttr-source}
\begin{tabular}{lrrrr}
\hline
\textbf{} & \textbf{Human} & \textbf{GPT} & \textbf{Llama} & \textbf{Qwen} \\
\hline
Total Words & 105,077 & 144,904 & 164,209 & 122,180 \\
Unique Words & 13,574 & 6,818 & 5,868 & 8,480 \\
TTR & 0.1292 & 0.0471 & 0.0357 & 0.0694 \\
\hline
\end{tabular}
\end{table}

The Table~\ref{tab:ttr-source} explains that individual human poets used more varied vocabulary, while AI-generated corpora shared more overlapping lexical patterns resembling the findings of Chen et al.\cite{22}. 

\subsubsection{Model-Specific Vocabulary Tendencie}\label{sec:5.1.2}
The human corpus is mixed with different types of vocabulary, while the three AI models show more concentrated tendencies. GPT focused on atmospheric vocabulary, Llama focused on emotional abstractions, and Qwen produced variations closest to human vocabulary. Thus, human poetry remains more diverse.

\begin{figure}[H]
    \centering
    \includegraphics[width=\columnwidth]{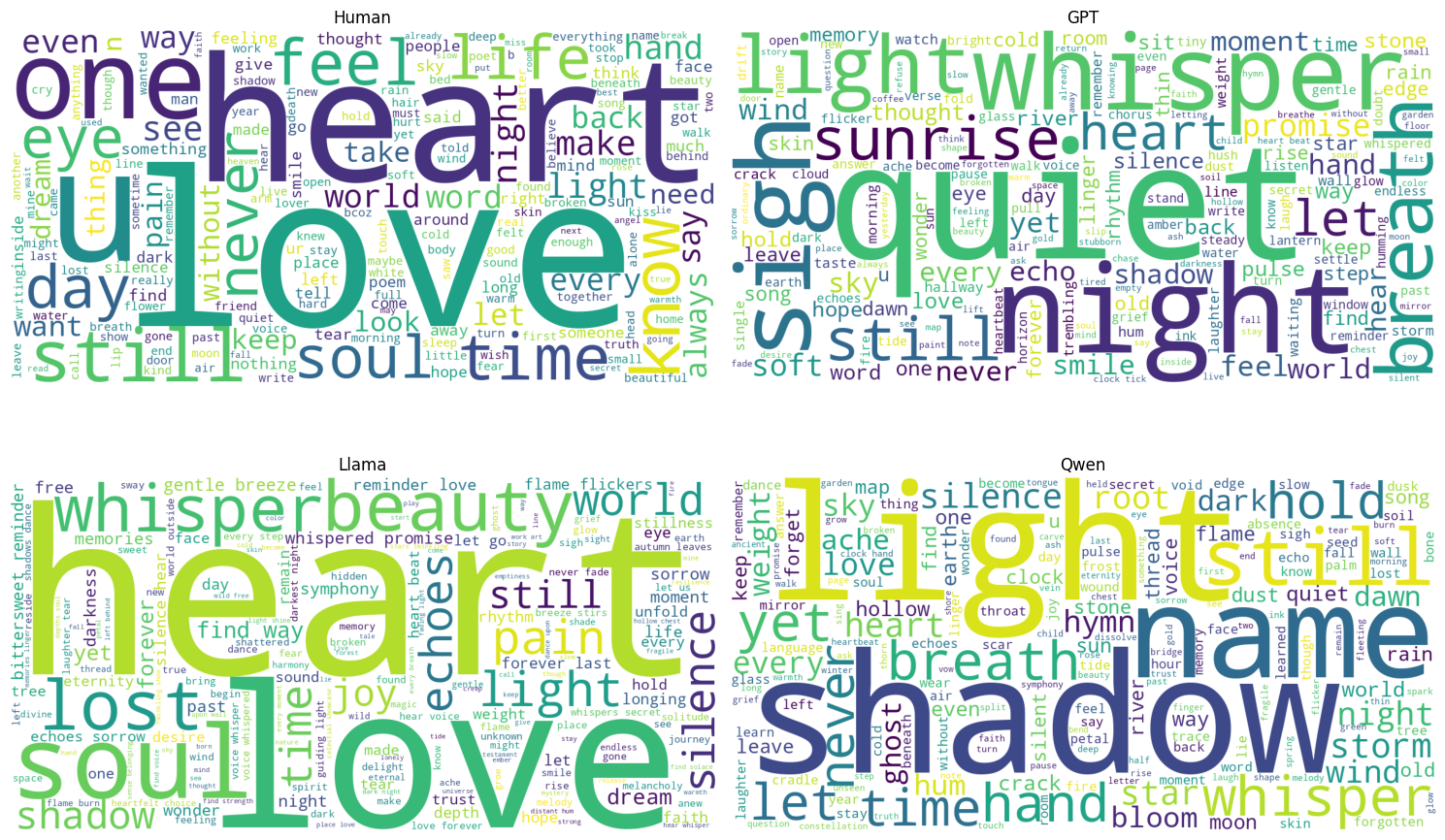}
    \caption{Word frequency distribution visualization}
    \label{fig:word-frequency-distribution}
\end{figure}

\subsubsection{Bigram and Trigram Feature Extraction}
The analysis also shows phrase-combination patterns. Table~\ref{tab:unique-ngram-counts} shows unique bigram and trigram counts per source.

\begin{table}[H]
\centering
\caption{Unique bigram and trigram counts by source}
\label{tab:unique-ngram-counts}
\begin{tabular}{lrr}
\toprule
\textbf{Source} & \textbf{Unique Bigrams} & \textbf{Unique Trigrams} \\
\midrule
Human & 87797 & 98987 \\
GPT-OSS-120B & 93433 & 135867 \\
Llama 3.3-70B & 77674 & 131682 \\
Qwen 3-32B & 95580 & 119651 \\
\bottomrule
\end{tabular}
\end{table}
\begin{table}[H]
    \centering
    \scriptsize
    \caption{Top 5 N-grams per Corpus.}
    \label{tab:top_ngrams}
    \renewcommand{\arraystretch}{1.12}
    \setlength{\tabcolsep}{1.5pt}
    
    \begin{tabular}{>{\raggedright\arraybackslash}p{0.13\columnwidth}>{\raggedright\arraybackslash}p{0.20\columnwidth}>{\raggedright\arraybackslash}p{0.20\columnwidth}>{\raggedright\arraybackslash}p{0.20\columnwidth}>{\raggedright\arraybackslash}p{0.20\columnwidth}}
        \toprule
        & \textbf{Human} & \textbf{GPT} & \textbf{Llama} & \textbf{Qwen} \\
        \midrule
        \textbf{Bigrams} 
        & 1. feels like \newline 2. love u \newline 3. miss lovely \newline 4. one day \newline 5. even though 
        & 1. feels like \newline 2. clock ticks \newline 3. heart beats \newline 4. night folds \newline 5. world feels 
        & 1. love that's \newline 2. heart that's \newline 3. find way \newline 4. gentle breeze \newline 5. whispered promise 
        & 1. clock hands \newline 2. thief steals \newline 3. second skin \newline 4. feels like \newline 5. hums hymn \\
        \midrule
        \textbf{Trigrams} 
        & 1. maybe maybe maybe \newline 2. feels like home \newline 3. loneliness looks like \newline 4. cyber cat tack \newline 5. life without u 
        & 1. steam curling like \newline 2. settle like dust \newline 3. wind carries scent \newline 4. echoes empty room \newline 5. clock ticks louder 
        & 1. like autumn leaves \newline 2. love that's lost \newline 3. gentle breeze stirs \newline 4. silence hear voice \newline 5. bittersweet reminder 
        & 1. like second skin \newline 2. clock without hands \newline 3. door left ajar \newline 4. clock hands drown \newline 5. shadows stretch like \\
        \bottomrule
    \end{tabular}
    
\end{table}

Human sample bigrams have colloquial phrases (Table~\ref{tab:top_ngrams}) that are mostly absent in AI poems. The three models use distinctive phrases: GPT uses atmospheric phrases, Llama uses emotional abstractions, and Qwen has more concrete and distinctive phrases. Such patterns can lead to misclassification of Qwen's poems as human-made. Finally, at the trigram level, repetition is rare.

\subsubsection{Theme-Based Analysis: Same Theme vs. Different Theme} Vocabulary patterns are influenced by theme frequency and were examined by dividing the poems into same-theme and different-theme groups. The same-theme group contained frequently generated words, while the different-theme group contained less frequent words. Although the theme distribution among the three models was mostly even, the same-theme group showed a limited set of recurring terms, whereas the different-theme group showed more vocabulary richness. The contrast clarified that same-theme poems had lexical repetition of poetic expressions, while different-theme poems introduced more varied word choices into the corpus.
\begin{figure}[H]
    \centering
    \includegraphics[width=\columnwidth]{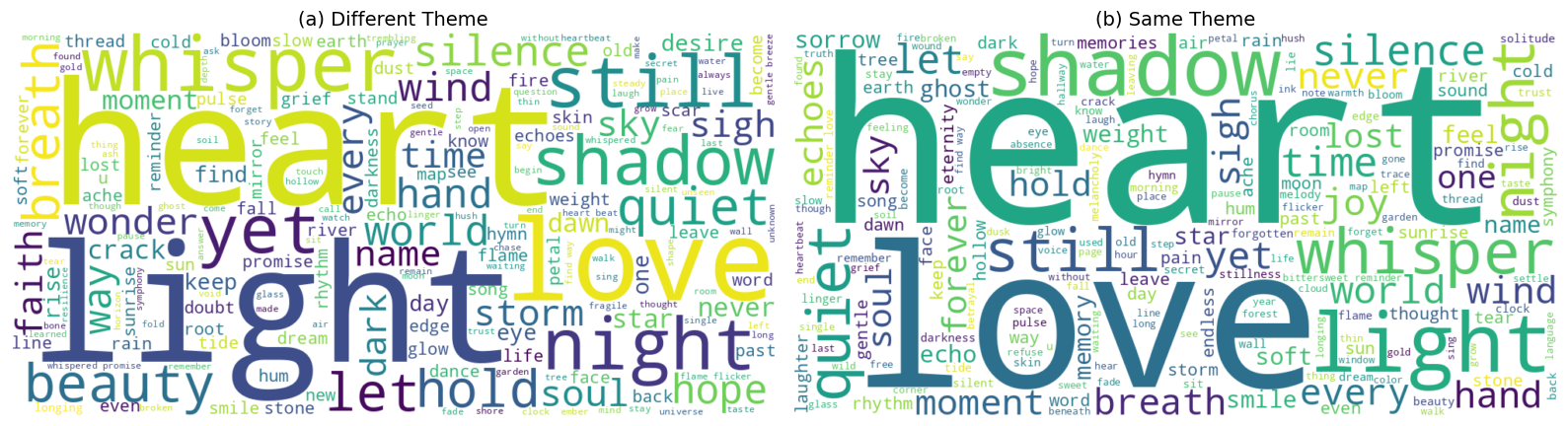}
    \caption{Theme-wise word cloud for same and different theme pairs}
    \label{fig:theme-wise-word-cloud-pairs}
\end{figure}

\subsubsection{Semantic Consistency Analysis}
Semantic consistency was measured by counting cases where a poem was assigned to a specific theme. This identifies whether the model stays focused on the assigned theme or followed the human sample semantics exactly, causing theme leakage.

\begin{figure}[H]
    \centering
    \includegraphics[width=\columnwidth]{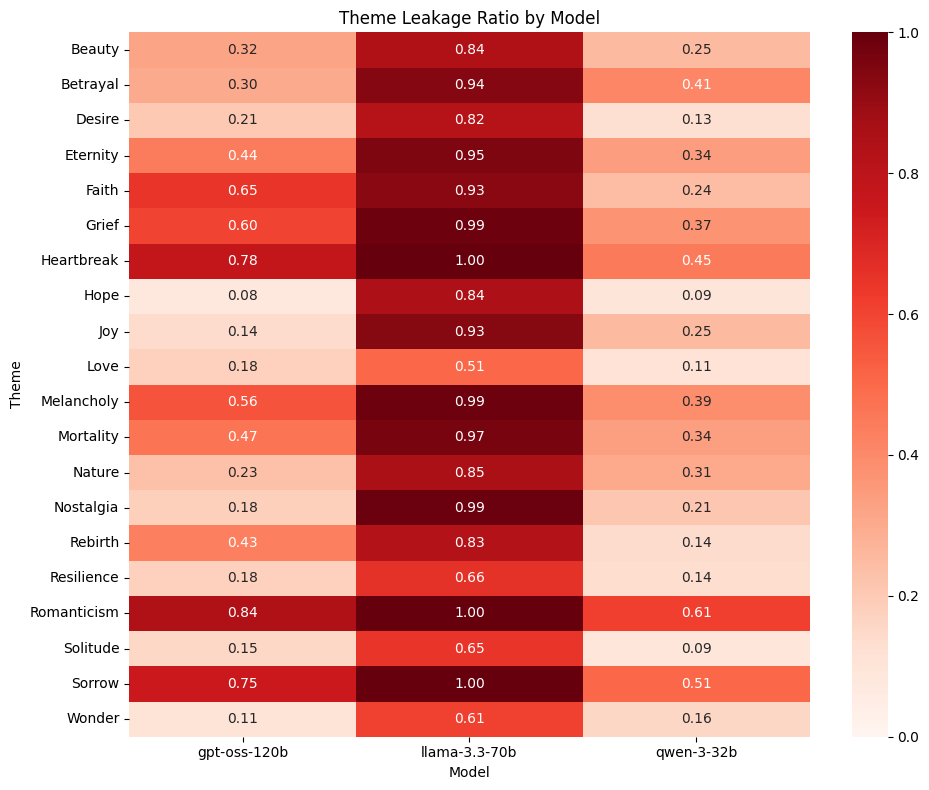}
    \caption{Semantic consistency analysis visualization}
    \label{fig:theme-word-repetition-analysis}
\end{figure}

The results indicate theme leakage, where models (Especially Llama) became more biased toward few-shot examples than toward the assigned theme.

\subsubsection{Prompt Repetition}

Prompt repetition shows direct repetition of assigned theme words such as beauty, love, and joy etc. in the generated poems. This was calculated by counting the frequency of assigned theme words appearing in poems of that theme. It measures thematic adherence where the model relies on the prompt word as a surface marker rather than developing the theme through poetic style.
\begin{figure}[H]
    \centering
    \includegraphics[width=\columnwidth]{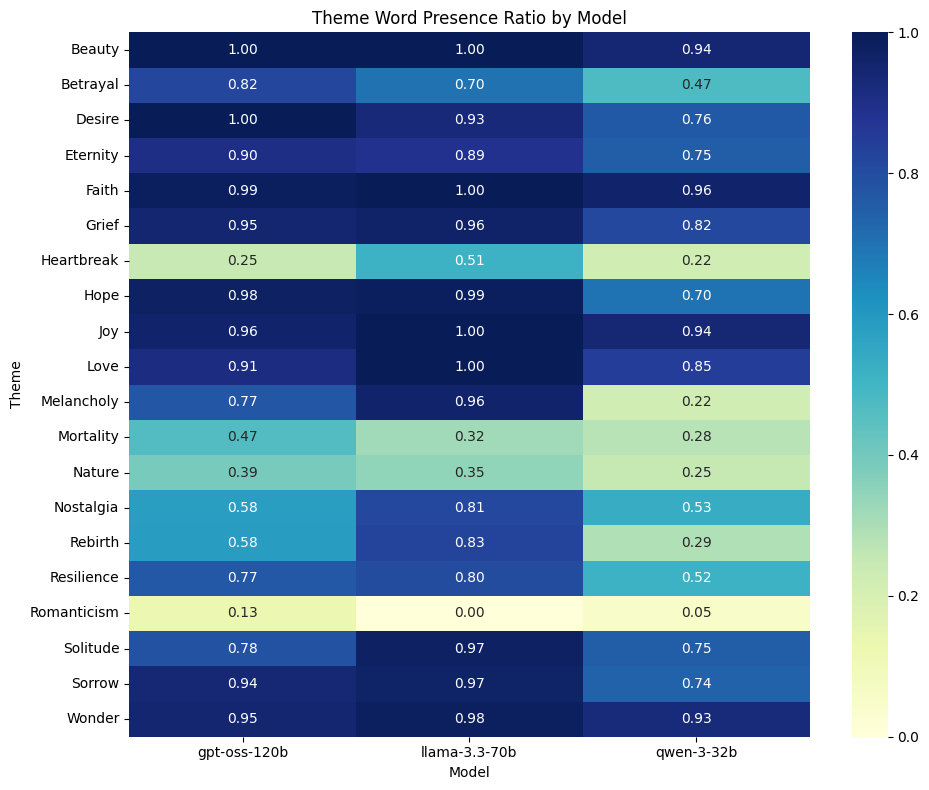}
     \caption{Theme word repetition analysis visualization}
    \label{fig:semantic-consistency-analysis}
\end{figure}

The results show that themes such as Beauty, Desire, Faith, Love, and Joy appeared quite often in generated poem bodies, but not enough to offset the few-shot bias.

\subsection{N-Gram Vocabulary Overlap Analysis: Human vs. AI}
The combined AI dataset shared 7514 unique words with the human corpus, representing 55.36\% of the human vocabulary and 62.99\% of the combined AI vocabulary. However, overlap dropped sharply at the bigram and trigram levels, showing that AI models may replicate human-like word choices while still producing distinct phrase combinations.

\begin{table}[H]
\centering
\caption{Full overlap metrics for Human versus AI combined}
\label{tab:human-ai-overlap}
\resizebox{\columnwidth}{!}{%
\begin{tabular}{lrrrrrr}
\hline
\textbf{Feature} & \textbf{Human} & \textbf{AI} & \textbf{Shared} & \textbf{Human Only} & \textbf{AI Only} & \textbf{Shared \% H/A} \\
\hline
Words & 13574 & 11928 & 7514 & 6060 & 4414 & 55.36 / 62.99 \\
Bigrams & 87797 & 240945 & 11605 & 76192 & 229340 & 13.22 / 4.82 \\
Trigrams & 98987 & 384398 & 660 & 98327 & 383738 & 0.67 / 0.17 \\
\hline
\end{tabular}}
\end{table}

Across GPT, Llama, and Qwen, bigram and trigram overlap values remained low, indicating that phrase combinations are both model-specific and human-specific. Qwen showed stronger lexical diversity, which aligns with its higher evasion tendency in later classification results.
 Finally, for the vocabulary overlap per model, the Table~\ref{tab:model-human-overlap} shows, at the word level, GPT has the highest share of its own vocabulary covered in human samples at 76.72\%, which is nearly three quarters of GPT's unique words. Llama follows at 74.52\%, while Qwen reaches 68.31\%. However, Qwen covers the largest share of human vocabulary at 42.68\%, compared with GPT at 38.54\% and Llama at 32.22\%, which is coherent with the TTR analysis.

\begin{table}[H]
\centering
\caption{Vocabulary and n-gram overlap between each AI model and Human}
\label{tab:model-human-overlap}
\resizebox{\columnwidth}{!}{%
\begin{tabular}{llrr}
\toprule
\textbf{Model} & \textbf{Feature} & \textbf{AI overlap with Human (\%)} & \textbf{Human overlap with AI (\%)} \\
\midrule
GPT & Words & 76.72 & 38.54 \\
GPT & Bigrams & 6.43 & 6.84 \\
GPT & Trigrams & 0.22 & 0.30 \\
Llama & Words & 74.52 & 32.22 \\
Llama & Bigrams & 6.70 & 5.93 \\
Llama & Trigrams & 0.23 & 0.31 \\
Qwen & Words & 68.31 & 42.68 \\
Qwen & Bigrams & 5.58 & 6.07 \\
Qwen & Trigrams & 0.14 & 0.17 \\
\bottomrule
\end{tabular}}
\end{table}

Across the three models, bigram and trigram values are quite low, making it clear that phrase combinations are model-specific and human-specific.

\subsection{Survey Result Analysis}\label{sec:5.3}
The experimental dataset includes a total of 320 survey evaluations collected across an allocated corpus of 180 unique poems. The corpus was divided into 9 distinct survey slots consisting of 20 poems each. The poems were evaluated independently by a panel of 16 distinct humans who have knowledge of English poetry and AI writings. Two slots were assigned as independent and unpaired consisting of 40 unique evaluations total and the remaining 7 slots consisting 140 unique poems were evaluated pairwise using a cross-dyad peer condition. The evaluation matrix consisted of 240 instances of AI-generated poems and 80 human-authored poetry. At the macro level, paired human evaluators converged on identical classification for only 35.71\% of the mutually reviewed corpus.\\ 

\begin{figure}[H]
    \centering
    \includegraphics[width=0.95\columnwidth]{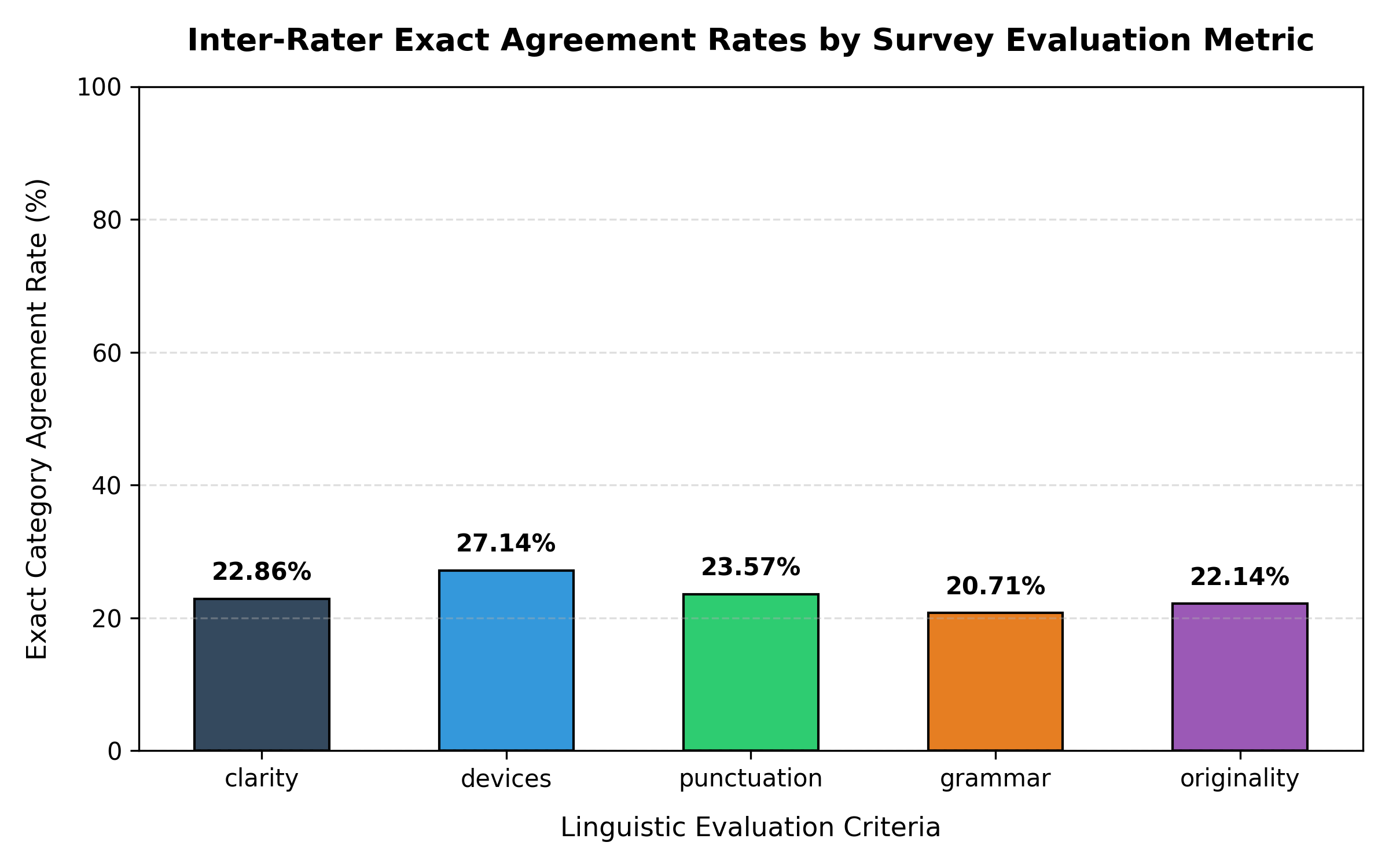}
    \caption{Inter-rater exact agreement rates by survey evaluation metric.}
    \label{fig:criterion-wise-agreement}
\end{figure}

 At the criterion level, agreement was even lower, as shown in Figure~\ref{fig:criterion-wise-agreement}: literary devices reached 27.14\%, punctuation 23.57\%, clarity 22.86\%, originality 22.14\%, and grammar 20.71\%. This indicates high perceptual subjectivity in human judgment.

\begin{figure}[!t]
    \centering
    \includegraphics[width=0.95\columnwidth]{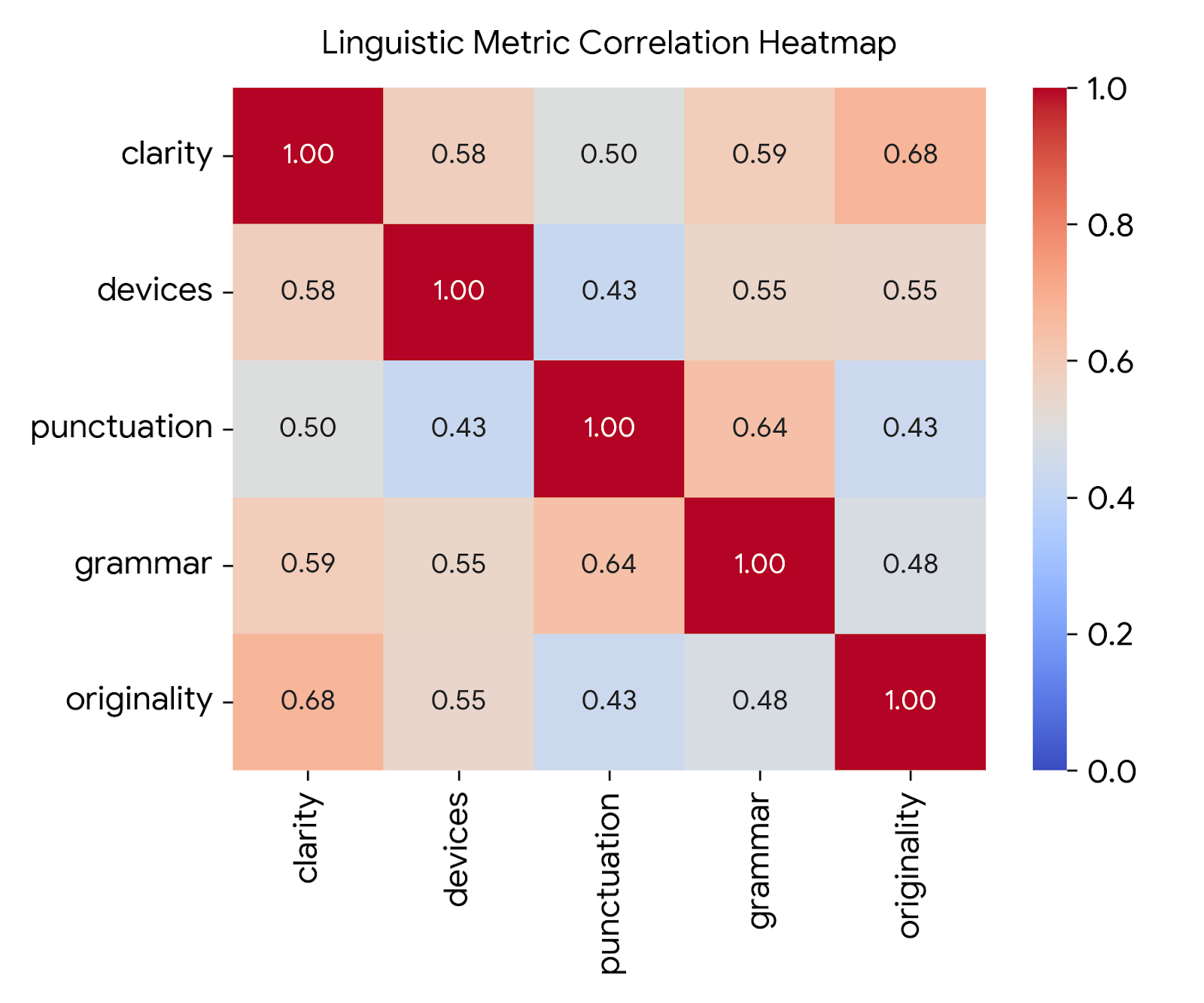}
    \caption{Linguistic metric correlation heatmap.}
    \label{fig:findings-systemic-blindspots}
\end{figure}

Our observational findings show that generative AI has attained absolute stylistic balance with human poets, revealing severe blindspots in human evaluation. This is characterized by a strong correlation ($r = 0.68$) between poem clarity and observed originality in Figure~\ref{fig:findings-systemic-blindspots}, which severely contributed to an 83.33\% human misclassification rate for human-authored poems. In addition, 30 AI-generated poems attained absolute camouflage with a 0.00\% detection rate, highlighting the complete breakdown of traditional creative limits between human and AI.

\begin{table}[H]
\centering
\caption{Human Classification Performance Report}
\label{tab:classification-report}
\resizebox{\columnwidth}{!}{%
\begin{tabular}{lcccc}
\hline
\textbf{Text Origin Class} & \textbf{Precision} & \textbf{Recall} & \textbf{F1-Score} & \textbf{Support} \\
\hline
AI-Generated & 0.7032 & 0.4542 & 0.5519 & 240 \\
Human-Authored & 0.2061 & 0.4250 & 0.2776 & 80 \\
\hline
\textbf{Global Accuracy} & & & 0.4469 & 320 \\
\textbf{Macro Average} & 0.4546 & 0.4396 & 0.4147 & 320 \\
\textbf{Weighted Average} & 0.5789 & 0.4469 & 0.4833 & 320 \\
\hline
\end{tabular}}
\end{table}

For the 240 true AI evaluations, human evaluators achieved a true positive classification rate of 45.43\% (110 items), while misidentifying 54.16\% (130 items) as human poems. For the 80 human-authored poems, evaluators correctly identified poem origin in only 42.50\% of cases (34 items), while misidentifying 57.50\% (46 items) as AI-generated text.

As shown in Table~\ref{tab:classification-report}, the overall human classification accuracy was 44.69\%.The imprecision profile (AI: 0.7032; Human: 0.2061) is primarily a result of imbalanced class support in the experimental corpus. The important metric for assessing human capability is recall, which is balanced yet suppressed across both AI recall at 45.42\% and human recall at 42.50\%. The severely low F1-score for human detection (0.2776) establishes that generative language models have attained stylistic consistency with human authors.

\subsection{Results of Gemma Classification}
 In zero-shot classification, Gemma 4 achieved stronger classification performance than the traditional detectors outpermoning the next best model (Log-Likelihood) by 3.5\% in \textbf{Weighted F1} score. 

\begin{table}[H]
\centering
\caption{Gemma 4 Classification Performance}
\label{tab:gemma-classification-performance}
\resizebox{\columnwidth}{!}{%
\begin{tabular}{lcccc}
\hline
\textbf{Evaluation Domain} & \textbf{Precision} & \textbf{Recall} & \textbf{F1-Score} & \textbf{Support} \\
\hline
AI Generated Poems & 0.933 & 0.933 & 0.933 & 4457 \\
Human Original Poems & 0.802 & 0.802 & 0.802 & 1513 \\
\hline
\textbf{Overall Accuracy} & \multicolumn{3}{c}{\textbf{0.900}} & \textbf{5970} \\
\textbf{Macro Average} & 0.868 & 0.868 & 0.868 & 5970 \\
\textbf{Weighted Average} & 0.900 & 0.900 & 0.900 & 5970 \\
\hline
\end{tabular}}
\end{table}
\begin{figure}[H]
    \centering
    \includegraphics[width=0.9\columnwidth]{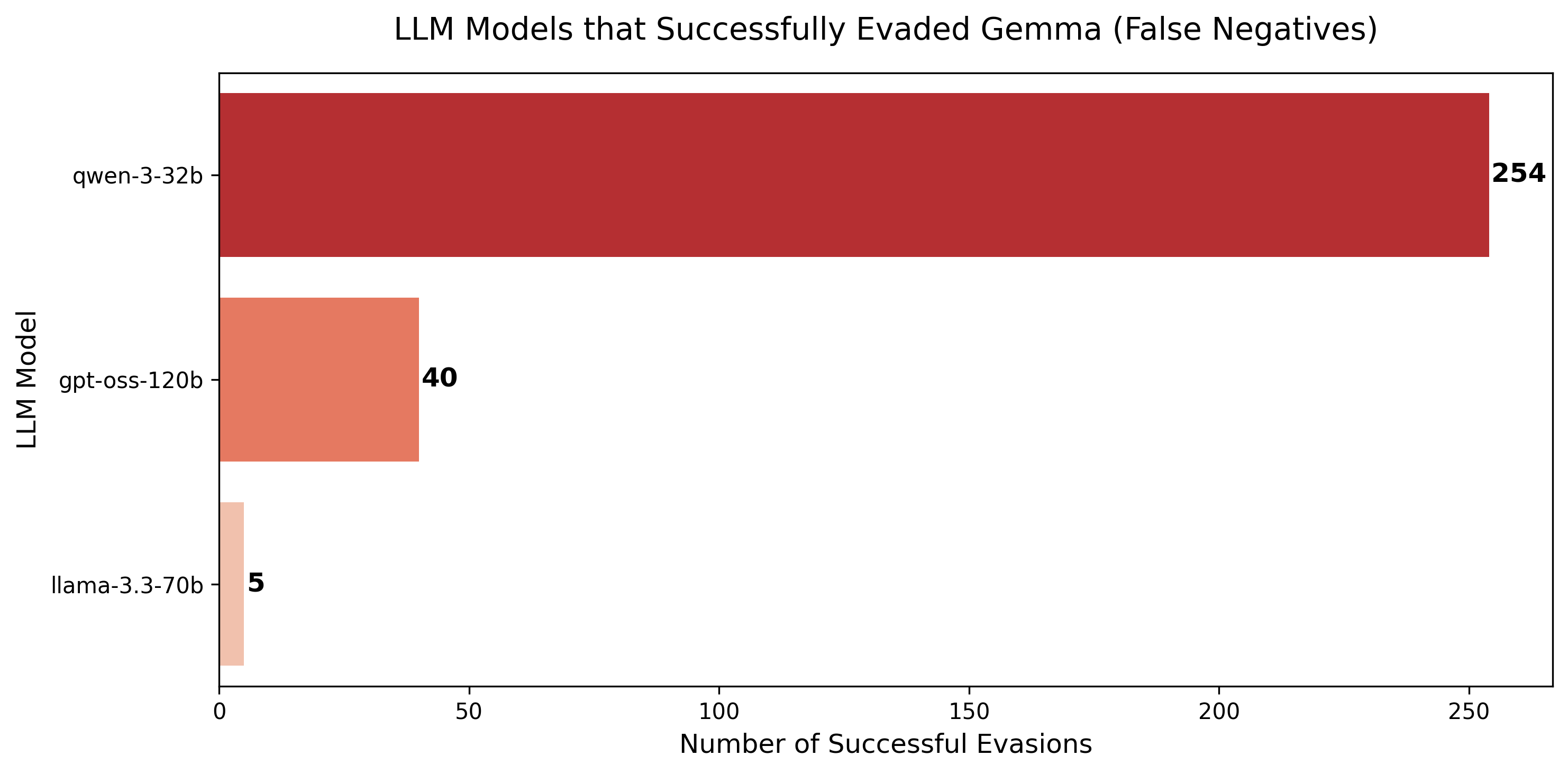}
    \caption{Number of False Negative classification produced by LLMs.}
    \label{fig:gemma_evasion_ranking.png}
\end{figure}

Qwen evaded Gemma the most (Fig.~\ref{fig:gemma_evasion_ranking.png}), which is consistent with the word diversity results. GPT-OSS-120B and Llama-3.3-70B were classified more efficiently, while Qwen's concrete and distinctive phrasing made it more difficult to detect.

\begin{table}[H]
\caption{Modal Criterion Scores for Correct Classifications, Positive (AI)}
\label{tab:criteria-mode-success}
\resizebox{\columnwidth}{!}{%
\begin{tabular}{lcc}
\toprule
 \textbf{Criteria} & \textbf{Most frequent Label for TP} & \textbf{Most frequent Label for TN } \\
\midrule
Poem Clarity & Unsure & Unsure \\
Literary Devices & Likely & Unlikely \\
Punctuation \& Spacing & Unlikely & Unlikely \\
Grammar \& Spelling & Unsure & Unsure \\
Originality & Likely & Unsure \\
\bottomrule
\end{tabular}}
\end{table}

\begin{table}[H]
\caption{Modal Criterion Scores for Prediction Errors, Positive (AI)}
\label{tab:criteria-mode-errors}
\resizebox{\columnwidth}{!}{%
\begin{tabular}{lcc}
\toprule
\textbf{Criteria} & \textbf{Most frequent Label for FP} & \textbf{Most frequent Label for FN} \\
\midrule
Poem Clarity & Unsure & Unsure \\
Literary Devices & Likely & Unsure \\
Punctuation \& Spacing & Unlikely & Unlikely \\
Grammar \& Spelling & Unsure & Unsure \\
Originality & Likely & Unsure \\
\bottomrule
\end{tabular}}
\end{table}
At the Table~\ref{tab:criteria-mode-success} for true positive classification, modal frequency indicated for the Criteria, Literary Devices and Originality Gemma frequently judged a poem to be AI and for the Criterion Punctuation and Spacing Gemma frequently judged a poem to be Human. Therefore,
\vspace{0.5em} 
\begin{itemize}
\item For TP predictions, \textbf{Literary Devices} and \textbf{Originality} was most helpful and \textbf{Punctuation and Spacing} was the least helpful. 
\item For true negative classification, \textbf{Literary Devices} and \textbf{Punctuation and Spacing} were the most helpful. 
\item Moreover, the Table~\ref{tab:criteria-mode-errors} indicates for false positive classifications, \textbf{Literary Devices} and \textbf{Originality} hindered Gemma and \textbf{Punctuation and Spacing} helped Gemma. 
\item Finally, for false negative classifications, \textbf{Punctuation and Spacing} only hindered Gemma. 
\end{itemize}. 

Hence, in both hard positive and hard negative classifications, the criterion \textbf{Literary Devices} had the most impact.
\vspace{0.5em}
\subsection{Results of the Traditional Detectors}
Among the traditional detectors (See Table~\ref{tab:detailed-classification}), Log-Likelihood \cite{35} performed best, but it was still less robust than Gemma 4 and Log Rank Ratio (LRR) \cite{34}, Binoculars \cite{33} performed second best and the third best respectively. Finally, Fast-DetectGPT \cite{32} performed the worst, contradicting the expectation that DetectGPT-derived approaches would be resilient against non-paraphrased AI-generated poetry.

\begin{table}[!t]
\centering
\caption{Detailed Classification Reports for AI Text Detectors}
\label{tab:detailed-classification}
\resizebox{\columnwidth}{!}{%
\begin{tabular}{lccccc}
\hline
\textbf{Detector / Domain} & \textbf{Precision} & \textbf{Recall} & \textbf{F1-Score} & \textbf{Support} & \textbf{Weighted F1} \\ 
\hline
Log-Likelihood AI & 0.911 & 0.907 & 0.909 & 4457 & \multirow{2}{*}{0.865} \\
Log-Likelihood Human & 0.730 & 0.738 & 0.734 & 1513 & \\
\hline
LRR AI & 0.882 & 0.843 & 0.862 & 4457 & \multirow{2}{*}{0.802} \\
LRR Human & 0.591 & 0.668 & 0.627 & 1513 & \\
\hline
Binoculars AI & 0.871 & 0.521 & 0.652 & 4457 & \multirow{2}{*}{0.610} \\
Binoculars Human & 0.354 & 0.773 & 0.485 & 1513 & \\
\hline
Fast-DetectGPT AI & 0.805 & 0.477 & 0.599 & 4457 & \multirow{2}{*}{0.552} \\
Fast-DetectGPT Human & 0.300 & 0.659 & 0.412 & 1513 & \\
\hline
\end{tabular}%
}
\end{table}

\subsection{Qualitative Attributes of GenAI and Human-Authored Poems}

Correctly classified GenAI poems often showed rigid rhyme schemes, symmetrical metaphors, generic sentimental phrasing, and predictable literary devices. However, some GenAI poems were misleading because they used non-cliché metaphors, colloquial contractions, irregular grammar, and human-like stylistic imperfections.

Human-authored poems were often identifiable through non-standard grammar, text-speak abbreviations, idiosyncratic metaphors, unconventional phrasing, and non-linear emotional arcs. However, some human poems were misclassified because they used cliché metaphors, predictable rhyme schemes, polished structures, and AI-like vocabulary. These findings show that both human-like GenAI attributes and AI-like human attributes contribute to classification errors. Detailed elaboration is presented at Table~\ref{tab:gemma_reasoning_summary}.

\begin{table}[!t]
\centering
\caption{Percentage of Gemma's classification reasoning factors for different scenarios}
\label{tab:gemma_reasoning_summary}
\vspace{3pt}
\scriptsize
\setlength{\tabcolsep}{3pt}
\renewcommand{\arraystretch}{1.12}
\resizebox{\columnwidth}{!}{%
\begin{tabular}{|>{\raggedright\arraybackslash}p{3.1cm}|>{\raggedright\arraybackslash}p{5.2cm}|>{\raggedleft\arraybackslash}p{1.3cm}|}
\hline
\textbf{Scenario} & \textbf{Factor} & \textbf{\% of Cases} \\
\hline
\textbf{AI poems (Qwen-3-32B)} misclassified as Human & Specific, grounded and visceral imagery & 24.8\% \\
\cline{2-3}
 & Generic and clich\'{e}d metaphors & 15.4\% \\
\cline{2-3}
 & Execution of general literary devices & 12.2\% \\
\cline{2-3}
 & Avoidance of AI clich\'{e}s and generic tropes & 11.4\% \\
\cline{2-3}
 & Emotional authenticity, vulnerability and depth  & 10.6\% \\
\cline{2-3}
 & Non-standard grammar, spelling and text-speak & 8.7\% \\
\hline
\textbf{AI poems (LLaMA-3.3-70B)} misclassified as Human & Non-standard grammar, spelling and text-speak & 80.0\% \\
\hline
\textbf{AI poems (GPT-OSS-120B)} misclassified as Human & Non-standard grammar, spelling and text-speak & 62.5\% \\
\cline{2-3}
 & Specific, grounded and visceral imagery & 5.0\% \\
\cline{2-3}
 & Execution of general literary devices & 5.0\% \\
\hline
\textbf{AI poems} correctly identified as AI & Rigid and predictable rhyme scheme & 36.7\% \\
\cline{2-3}
 & Generic and clich\'{e}d metaphors & 10.1\% \\
\cline{2-3}
 & Repetitive, formulaic and mechanical structure & 8.0\% \\
\cline{2-3}
 & Avoidance of AI clich\'{e}s and generic tropes & 7.9\% \\
\cline{2-3}
 & Emotional authenticity, vulnerability and depth & 6.2\% \\
\cline{2-3}
 & Execution of general literary devices & 5.6\% \\
\hline
\textbf{Human poems} correctly identified as Human & Non-standard grammar, spelling and text-speak & 47.0\% \\
\cline{2-3}
 & Grounded and visceral imagery & 10.0\% \\
\cline{2-3}
 & Rigid and predictable rhyme scheme & 6.3\% \\
\cline{2-3}
 & Generic and clich\'{e}d metaphors & 4.9\% \\
\cline{2-3}
 & Idiosyncratic \& unconventional metaphors & 3.6\% \\
\cline{2-3}
 & Execution of general literary devices & 3.5\% \\
\hline
\textbf{Human poems} misclassified as AI & Rigid, predictable rhyme scheme & 30.1\% \\
\cline{2-3}
 & Generic and clich\'{e}d metaphors & 26.8\% \\
\cline{2-3}
 & Execution of general literary devices & 8.7\% \\
\cline{2-3}
 & Emotional authenticity, vulnerability and depth & 5.7\% \\
\cline{2-3}
 & Avoidance of AI clich\'{e}s and generic tropes & 5.4\% \\
\cline{2-3}
 & Repetitive, formulaic and mechanical structure & 5.0\% \\
\hline
\end{tabular}}
\end{table}

\section{Conclusion}
The Conclusion is divided into 3 subsections, A. Findings B. Limitations and C. Future Work. The further elaboration of the aforementioned subsections are stated as,

\subsection{Findings}
The key findings of this research remain,
\begin{itemize}
    \item \textbf{Classification Performance}: Gemma 4 performed best by the holding best \textbf{Weighted F1 Average}, exceeding in every classification criteria, albiet by a small margin. This performance suggests that Gemma 4-31B can be light weight and robust alternative to the current for-profit detection tools. Additionally, Fast-DetectGPT's poor performance contradicts the findings of our literature review. As Fast-DetectGPT is derived from DetectGPT \cite{5} hence, it was supposed to robust against our non-paraphrased Poem dataset but amongst all the AI based detectors, it acheieved the lowest \textbf{Weighted F1 Average} score of 0.552. Finally, human classification was also poor with 0.70 precision and 0.45 recall in AI generated portion and only 0.20  precision and 0.42 recall score in human-written part, affirming the indistinguishability AI poems to human poems \cite{9}. 

    \item \textbf{Human-like and Non-Human-like Attributes}: In our reserach we were able to find the Human-like and Non-Human-like attributes (Defining Attributes) and contrastingly we were also able to find Misleading attributes. In most cases, the rigid and the cliché nature of GenAI compostion aided significantly in detection. Seemingly, the opppsite is also true. GenAI composition with unique usage of literary devices (Mainly, Metaphors and Imageries) aided in evasion. Qwen leveraged this misleading attribute to evade Gemma 4 the most. Similarly, usage of unconventional metaphors, unique grammartical structure and spelling were also part of prominent misleading attributes (See Example:~\ref{Ex: 1}). Finally, human writing in predictable rhyme and generic patterns is also in the risk of misclassification (See Example: ~\ref{Ex: 2}). Hence, the misclassification via misleading attributes needs mending.

    \item \textbf{Biased towards Few-shot examples over User Prompt}: In generating, the 3 LLMs had the tendency to be biased towards few-shot examples over user prompts. This assumption is supported by the semantic consistency analysis. In the generation loop, themes were kept independent to what the authors writing style to ensure better lexical diversity and originality but the disconnection between the user given poetry theme and the generated poem's theme made such separation dificult achieve. Similar phenomena also occured when in Style Guide, it was explicitly mentioned not to use "em" dashes (Subsection~\ref{sec:4.1.1.2.1} ) but human authors' usage of "em" dashes in few-shot example overruled the style based prompt command. Therefore, we had to extensively clean the "em" dashes in the Data Pre-processing section.

    \item \textbf{Lack of Word Diversity }: Biased towards fewshot blocks can be a reason behind such excessive word repetition as the few-shots examples were picked randomly for each generation based on the author. Hence, it is possible that same few-shots examples were picked excessive amount of times by Python's random library and as the models exhibit more biased towards few-shots examples over prompt based themes so such lack of theme adherence decreased word diversity and forced repetitive generations. 

\end{itemize}

\subsection{Limitations} 
The limitations of this study include inadequate cleaning of the dataset. Even after pre-processing, the potential ambiguity in the constructed dataset cannot be fully ruled out. Additionally. during the generation of the final dataset, certain uncertainties and biasness in the universal style guide were noticed. For example, the style guide instructed not to add any em dashes along with other restrictions. However, analysis of the AI-generated poetries showed that some models still generated poems containing em dashes which was later removed through preprocessing. Moreover, while conducting the survey, another input called 'Overall Label' should have been taken along with other five evaluation criteria. Additionally, as our research is based on English poetries, the response from native English speakers should have been taken as well during the survey. Furthermore, humanizing GenAI texts is crucial to evade the detection tools. Yet, we did not use any paraphrasers or humanizers in our dataset as implementing humanizing framework in poetry is extremely difficult without damaging the poem integrity by altering the structure and the context of the poem. 

\subsection{Future Work}
Future work remains to train Gemma 4-31B with hard negative examples (i.e., poems with misleading attribute). Training the model only on such attributes will not only reduce the training margin needed but also gives us a clearer picture of which framework is truly better while detecting GenAI poetry. Finally, Qwen 3-32B's high evasiveness needs further investigation via adding new detectors in the classification pipeline.

\appendices
\section{Supplementary Dataset Details}
\begingroup
\setlength{\parskip}{0.1em}
\setlength{\parindent}{0pt}
\renewcommand{\baselinestretch}{0.9}\selectfont

\subsection{Usage of non-cliché metaphors and imagery by Qwen 3-32B}\label{Ex: 1}
\textit{Silent Room}\\

A coat hangs on the back of the chair\\
where she once sat, knitting silence\\
into the air. The clock stopped\\
when she left. Dust motes drift\\
like forgotten seconds.\\
I open her jar of perfume\\
a breath of bergamot, then nothing.\\
The teacup cools beside the book\\
she never finished.\\
Outside, the wind wears her laugh\\
like a borrowed coat. I close the door\\
and let the dark collect itself\\
into a shape that fits the empty.

\subsection{Predictable rhyme scheme and generic writing by a human author}\label{Ex: 2}
I really miss what we had, not how it all fell apart,\\
Not the final goodbye or the ache in my heart,\\

I do not wanna touch the ending, I do not want to rewrite the pain,\\
I honor the love that still leaves a soft stain,\\
We were real in those moments, the laughter, the light,\\
The way being seen felt so warm, so right,\\
No damage, no anger, no need to pretend,\\
Just proof that our love did not break, it just bent,\\
We found each other once, and that still feels true,\\
Like a mark on my soul shaped a little like you,\\
Some loves are not forever, but that does not mean,\\
They were not something holy, or deeply unseen,\\
So I carry the knowing, quiet and kind,\\
That love once found us still lives in my heart and mind,\\
And maybe in another life, in another way,\\
What we were together gets the chance to stay.

\endgroup
\section{Example of Final Generated Prompt}
\label{app:prompt_example}

Example of the fully constructed prompt for the author, \textit{Garrie Grant} with the theme \textit{Nature}:\par
\vspace{0.6em}
\begingroup
\normalsize
\setlength{\parskip}{0pt}
\setlength{\parindent}{0pt}
\renewcommand{\baselinestretch}{0.92}\selectfont
\noindent STYLE GUIDE:\\
1. TITLE: Provide a meaningful, original TITLE at the very top of the poem.\\
2. CONSTRAINTS: Only English language poems are allowed. Do NOT use em dashes (—).\\
3. ORIGINALITY: Do NOT plagiarize or directly copy the provided author's\\
work. Absorb their stylistic voice, but generate a 100\% original poem\\
based on the given theme.\\[0.35em]
Here are 2 examples of Garrie Grant's writing style. Use these ONLY to\\
understand their voice, rhythm, and formatting. Do NOT copy the content.\\
Write a completely new poem based on the assigned theme.\\[0.35em]
--- EXAMPLE 1 ---\\
TITLE: Drinking without occasion on a Saturday night in March.\\
POEM:\\
The sound of her voice like...\\
The street lights reflecting on a perfect...\\
--- END EXAMPLE ---\\[0.6em]
--- EXAMPLE 2 ---\\
TITLE: Afterglow\\
POEM:\\
Where did all the....\\
This darkness has no...\\
--- END EXAMPLE ---\\[0.35em]
Generate a brand new, original poem based on the following theme.\\
THEME: Nature
\endgroup

\section*{Acknowledgment}
The authors thank the participating poets and survey respondents for their contributions to the dataset and evaluation process.

\end{document}